\documentclass[11pt,a4paper]{article}
\usepackage[hyperref]{emnlp2020}
\usepackage{url}
\usepackage{times}
\usepackage{latexsym}

\usepackage{microtype}
\usepackage{booktabs}
\usepackage{textcomp}
\usepackage{amssymb}
\usepackage{graphicx}
\usepackage{tabulary}
\usepackage{pifont}
\newcommand{\cmark}{\ding{51}}%
\newcommand{\xmark}{\ding{55}}%

\usepackage{comment}
\usepackage{soul}

\usepackage[para]{footmisc}

\usepackage{adjustbox}
\usepackage{array}
\usepackage{xspace}
\newcolumntype{R}[2]{%
    >{\adjustbox{angle=#1,lap=\width-(#2)}\bgroup}%
    l%
    <{\egroup}%
}
\newcommand*\rot{\multicolumn{1}{R{75}{1em}}}

\newcommand{\ex}[1]{\textit{#1}\xspace}

\aclfinalcopy 

\title{COVID-SEE: Scientific Evidence Explorer\\ for COVID-19 Related Research}

\author{Karin Verspoor, Simon Suster, Yulia Otmakhova, Shevon Mendis,\\
  \textbf{Zenan Zhai, Biaoyan Fang, Jey Han Lau, Timothy Baldwin}\\
  The University of Melbourne  \\
  \textbf{Antonio Jimeno Yepes, David Martinez}\\
  IBM Research Australia \\
  contact: \texttt{karin.verspoor@unimelb.edu.au}   }

\date{}

\begin{document}
\maketitle

\begin{abstract}
 We present COVID-SEE, a system for medical literature discovery based on the concept of information exploration, which builds on several distinct text analysis and natural language processing methods to structure and organise information in publications, and augments search by providing a visual overview supporting exploration of a collection to identify key articles of interest.  We developed this system over COVID-19 literature to help medical professionals and researchers explore the literature evidence, and improve findability of relevant information. COVID-SEE is available at \href{http://covid-see.cis.unimelb.edu.au/}{covid-see.cis.unimelb.edu.au}.

\end{abstract}

\section{Introduction}


The outbreak of Coronavirus disease 2019 (COVID-19) has led to a rapid and proactive response from medical and AI communities worldwide. In information retrieval and natural language processing, efforts have concentrated on building datasets and tools for efficiently managing the growing literature on COVID-19, historical coronaviruses, and other related diseases \citep{Hutson2020}. One of the main stimuli for the creation of new technology was the release of the COVID-19 Open Research Dataset (CORD-19) \citep{wang-lo-2020-cord19}, a regularly updated collection of coronavirus-related publications and preprints. But while many tools have emerged for the purposes of article retrieval and question answering,\footnote{A non-exhaustive list:\ \tiny{\href{http://cord19.vespa.ai}{cord19.vespa.ai}, \href{http://discovid.ai}{discovid.ai}, \href{http://covidex.ai}{covidex.ai}, \href{http://cord19.aws}{cord19.aws}, \href{http://covid19.mendel.ai}{covid19.mendel.ai}, \href{http://covidscholar.org}{covidscholar.org}, \href{http://covidseer.ist.psu.edu}{covidseer.ist.psu.edu}, \href{http://covidask.korea.ac.kr}{covidask.korea.ac.kr}, \href{http://covid19-research-explorer.appspot.com}{covid19-research-explorer.appspot.com}}.}
relatively less work has been devoted to designing systems that go beyond returning a list of (relevant) documents, and that attempt to leverage domain knowledge to organise and present information found within the literature. With our work, we aim to fill this gap by making available a web application that combines a search engine for COVID-19-related literature with different visualisation techniques, that make it easy to see and explore salient topics, concepts and concept relations in the documents.

We build on observations about the relevance of \textit{exploratory search}, and the need to combine
 learning and investigation with direct retrieval  
for information seeking~\cite{marchioni2006}, specifically designing a tool to support the health information seeking behaviour observed by~\citet{pang2015} of alternating between exploratory and focused search. We further draw on the insights from research in information visualisation that demonstrate the value of multiple coordinated views of documents, with a specific emphasis on visually illustrating connections between entities~\cite{stasko2008,gorg2010visualization}. 


We designed our tool to support medical experts and other researchers by way of literature discovery. A typical usage scenario in COVID-SEE begins with a textual query over CORD-19, providing the researcher with: (i) a list of \textit{retrieved documents}, and (ii) a \textit{visualisation dashboard}. The user can then select the documents of interest and save them into a \textit{briefcase} for later use. Once the user has populated the briefcase with relevant articles, it is possible to set the briefcase as the current active collection, and further explore those articles.

\begin{figure*}[t]
    \centering
    \includegraphics[width=16cm]{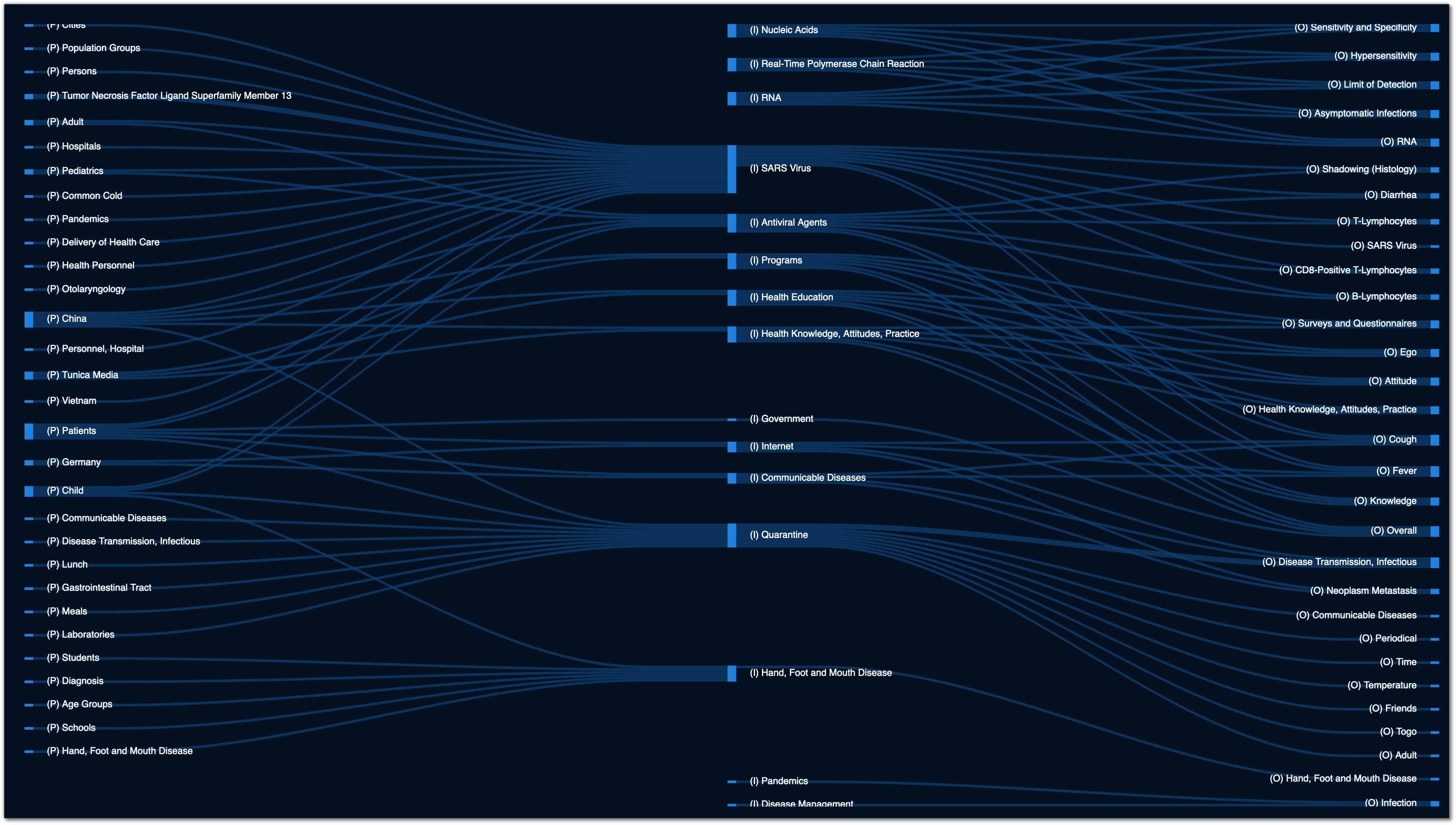}
    \caption{Sankey diagram of PICO concepts and relations for 58 articles retrieved for the query \ex{incubation period of COVID-19}. Links between concepts can be selected to reveal papers that contain those relations. (See Fig.~\ref{fig:inter-out-select})}
    \label{fig:sankey}
\end{figure*}
\begin{figure*}[t]
    \centering
    \includegraphics[width=16cm]{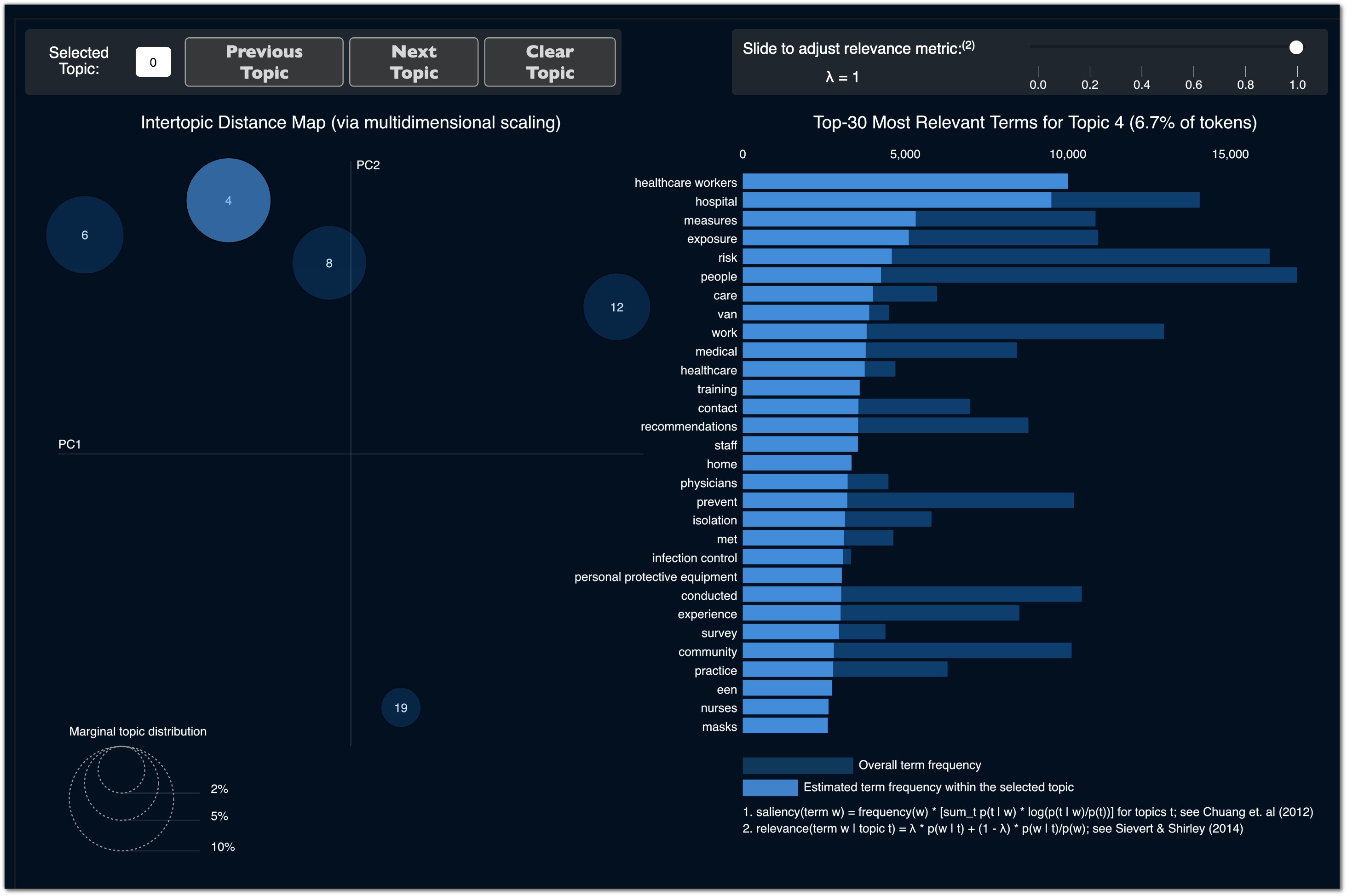}
    \caption{Topic visualisation for the 58 articles related to the query \ex{incubation period of COVID-19}.}
    \label{fig:topics}
\end{figure*}
\begin{figure}[t]
    \centering
    \includegraphics[width=8cm]{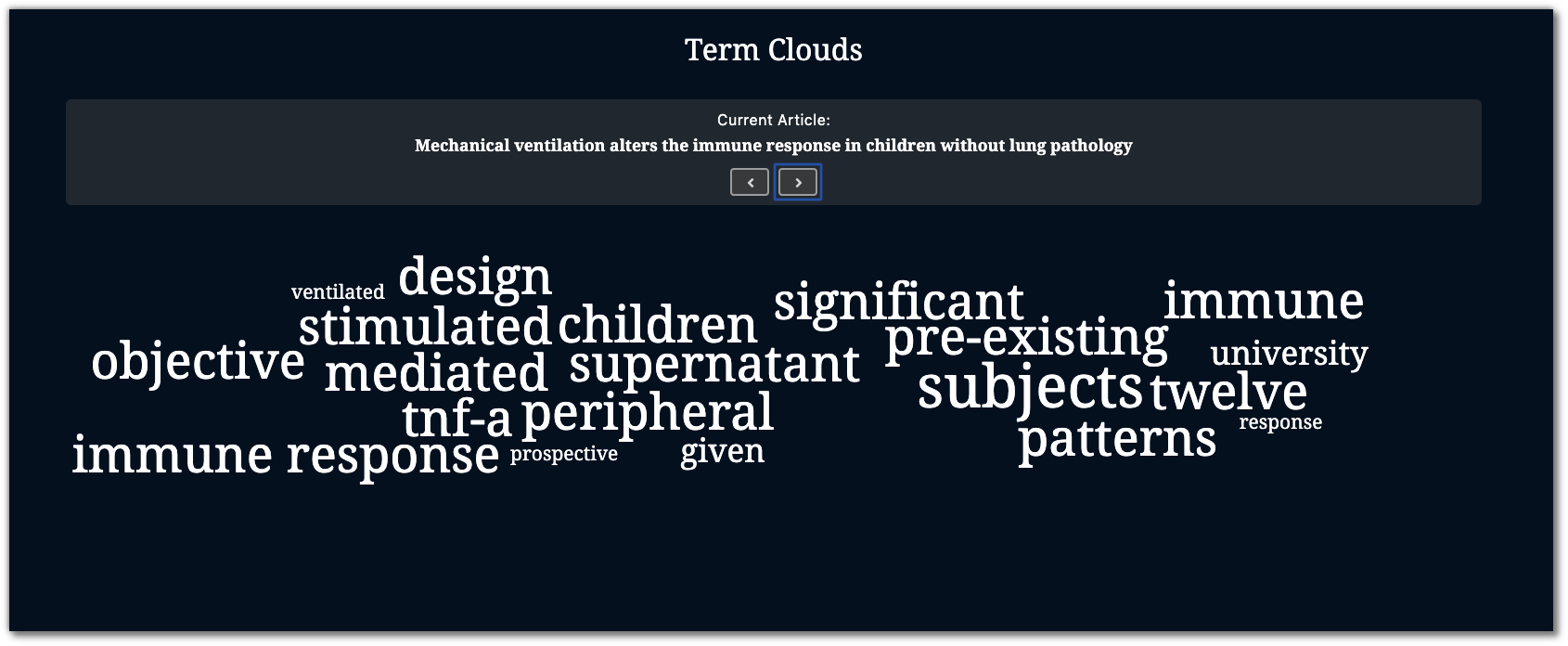}
    \caption{Concept cloud view of a single document.}
    \label{fig:cloud}
\end{figure}

In both the retrieval and visualisation components, we adopt several well-established NLP techniques. Article retrieval is powered by an existing neural search engine specifically developed for the CORD-19 dataset \cite{zhang2020rapidly}. The results are shown as a list of document hits with metadata about the articles. Separately, the dashboard represents the current active collection with three distinct \textit{interactive views}. The first is a \textbf{relational concept view} in which we employ Sankey diagrams to organise the medical concepts found in the articles according to broad, clinically-relevant PICO categories \citep{richardson1995well}, such as population, intervention and outcome. In this view, more salient relations carry more weight, and once a relation is clicked, the corresponding articles are revealed. We use an example based on the query \ex{incubation period of COVID-19} to illustrate this functionality (Fig.~\ref{fig:sankey}). 
The second, \textbf{topic view} (Fig.~\ref{fig:topics}), is more thematic and shows representative topics for the current collection. For this component, we trained a global topic model on medical concepts extracted from CORD-19. Our third component is a \textbf{concept cloud view} (Fig.~\ref{fig:cloud}), showing the most salient concepts for each active document.

\begin{figure}[t]
    \centering
    \includegraphics[width=8cm]{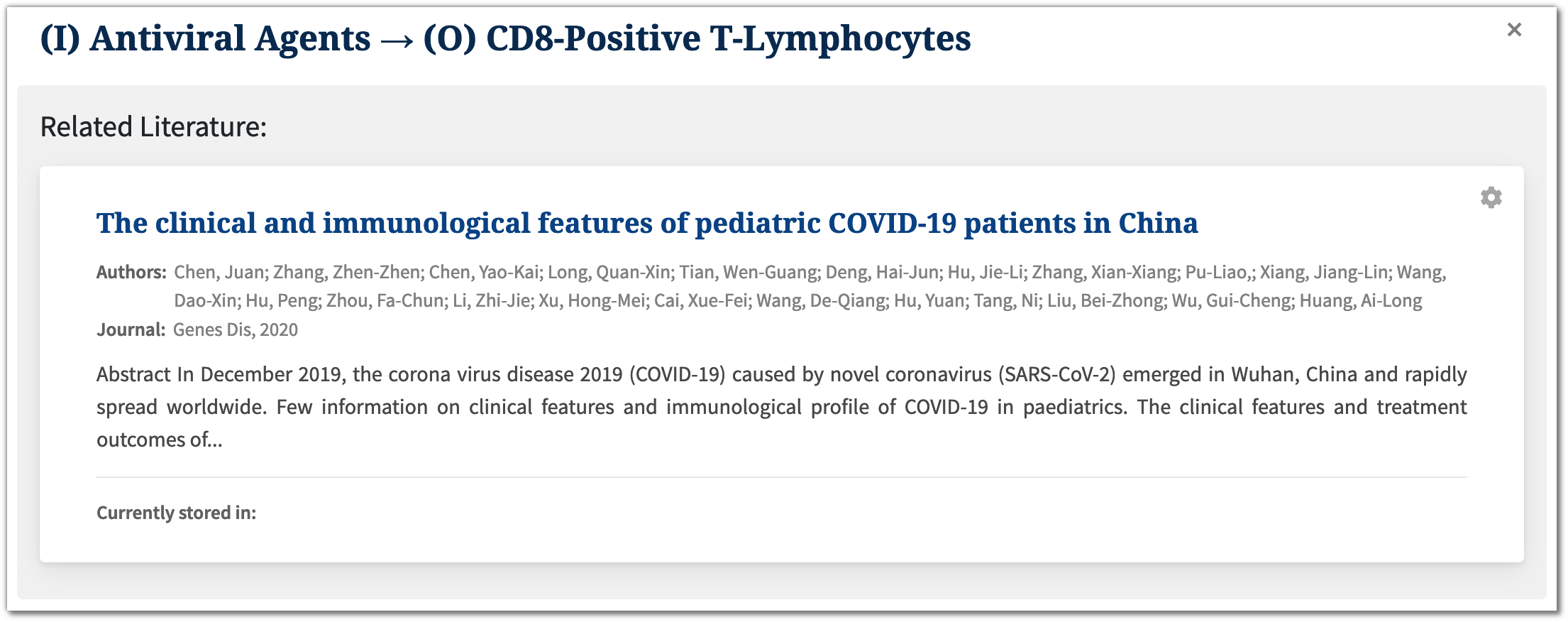}
    \caption{Selection of an Intervention-Outcome link reveals a paper containing the specific concept relations.}
    \label{fig:inter-out-select}
\end{figure}

\section{Data}
CORD-19 is currently the most extensive coronavirus literature corpus publicly available \citep{wang-lo-2020-cord19}. The dataset contains all COVID-19 and coronavirus-related research (e.g.\ SARS, MERS, etc.) from different sources, including PubMed's PMC open access corpus, research articles from a corpus maintained by the WHO, and bioRxiv and medRxiv pre-prints. The corpus is  updated daily and consisted of more than 130,000 documents as of 6 June. Together with the release of the CORD dataset, a Kaggle challenge\footnote{\tiny{\href{https://www.kaggle.com/allen-institute-for-ai/CORD-19-research-challenge}{kaggle.com/allen-institute-for-ai/CORD-19-research-challenge}}}
 took place in which we also participated.

\section{System overview}

\subsection{Information retrieval}
After submitting a query, a list of retrieved documents is shown. Each document entry can be expanded to display the entirety of the abstract as well as metadata (authors, journal, source, year, license). The user can also filter by criteria such as year and source. Articles can be selected and added to the user's briefcase. This briefcase represents the set of documents that the user wishes to keep track of, and can be visualised, versioned, and exported. 

We make use of an existing information retrieval system for the CORD-19 dataset, available at \href{https://covidex.ai/}{covidex.ai} \citep{zhang2020rapidly}. This provides us with a well-performing retrieval system---by our analysis, in the top 30\% of submissions to the TREC-COVID shared task \citep{roberts-etal-2020-trec}. 

\subsection{Relational concept view}

Given a collection of retrieved documents, the user can directly proceed to the dashboard to explore the collection. A key view over the collection is via a Sankey diagram frame, which organises the identified medical concepts into PICO categories (Population, Intervention/Comparator, Outcome), and shows how they interact.  The diagram displays relations between pairs of medical concepts, where the strength of a relation corresponds to the number of supporting abstracts in which that concept pair is attested. To obtain the PICO concepts, we follow a two-step procedure, as described below.

\begin{table}[t!]
\centering
\small
\begin{tabular}{r c c c}
\toprule
 & P & R & F1 \\
 \cmidrule(l){1-4}
All & 0.73 & 0.65 & 0.69 \\
Population & 0.78 & 0.79 & 0.78 \\
Intervention & 0.57 & 0.64 & 0.60 \\
Outcome & 0.80 & 0.58 & 0.67 \\
\bottomrule
\end{tabular}
\caption{PICO labelling results on EBM-NLP.}
\label{tab:pio_results}
\end{table}

\begin{table*}[t]
\centering
\small
\begin{tabular}{p{5cm} l p{6cm}}
\toprule
 PICO snippet & PICO category & MeSH terms \\
\cmidrule(l){1-3}
\ex{patients presenting with RTI} & Population & Patients; Respiratory Tract Infections \\
\ex{Mass vaccination campaigns with parenteral vaccines} & Intervention & Immunization Programs; Vaccines; Parenteral Nutrition; Macrophage Activation Syndrome \\
\ex{cumulative COVID-19-related hospitalization and death rates} & Outcome & Hospitalization; Mortality \\
\bottomrule
\end{tabular}
\caption{Examples of extracted PICO textual spans and MeSH terms found in them. The PICO concepts we use are the MeSH terms, typed by their respective PICO category (e.g.\ Vaccines+Intervention).}
\label{tab:pio_examples}
\end{table*}

\paragraph{PICO span identification}

We train a BiLSTM-CRF model \citep{lample-etal-2016-neural} on the EBM-NLP dataset \citep{nye-etal-2018-corpus}, which contains $\sim$5,000 MEDLINE\footnote{\tiny{\href{https://www.nlm.nih.gov/bsd/pmresources.html}{www.nlm.nih.gov/bsd/pmresources.html}}} abstracts of reports of randomised clinical trials annotated with textual spans that describe the PICO elements. As pretrained word representations for the model, we use 200-dimensional word2vec embeddings induced on PubMed abstracts and MEDLINE articles \citep{hakala-etal-2016-syntactic}. We observe a micro-averaged F1 score of 0.69 on the EBM-NLP test set, and report the full results in Table~\ref{tab:pio_results}; these are comparable to those reported in \citet{nye-etal-2018-corpus}. For our purposes, getting the textual span exactly correct is less important, as our final representations are not the snippets but the medical concepts that occur inside them. The PICO identification step is applied to all abstracts of the CORD-19 dataset.

\paragraph{Recognition of PICO concepts} In this step, we recognise medical concepts that occur within the identified PICO spans. This is part of a larger concept recognition procedure that we describe in more detail in Section~\ref{sec:concepts}. Here, we take the terms corresponding to Medical Subject Headings (MeSH),\footnote{\tiny{\href{https://www.ncbi.nlm.nih.gov/mesh}{ncbi.nlm.nih.gov/mesh}}} which is a structured vocabulary maintained by the National Library of Medicine. We use MeSH due to its transparent hierarchical structure, making it easy to control term granularity. Additionally, it is also an established way to represent keywords found in medical articles indexed by PubMed. A few examples of extracted PICO concepts are shown in Table~\ref{tab:pio_examples}.

\paragraph{Establishing relations}
The output of the previous two steps is the extracted MeSH terms for each article, typed by their respective PICO category. In the Sankey diagram, we then display pairwise relations between Population--Intervention and Intervention--Outcome concepts. Two concepts are related when they are found in the same abstract. We also include the strength of the connection, corresponding to the absolute frequency of that concept pair in the entire dataset. The user can select a concept relation link of interest, browse the relevant articles containing that relation, and save them to the briefcase for further exploration.



The rendering of the diagram is dynamic. This means that whenever a document collection selected by the user changes, the diagram is updated. In addition, the Sankey view is interactive, allowing the user to click on any relation to highlight the relevant part of the diagram and show the supporting documents.
A planned future release of our tool will support \textit{semantic search}, which will enable the use of PICO-typed MeSH terms as search criteria.

\subsection{Medical concept recognition}
\label{sec:concepts}
To allow for more conceptual representation of the documents, we transform them into an unordered set of Unified Medical Language System (UMLS: \citet{lindberg1993unified}) concepts. These concepts are extracted using MetaMap\footnote{\tiny{\href{http://metamap.nlm.nih.gov}{metamap.nlm.nih.gov}}} from the document's abstract, or, if it is absent, from the first two paragraphs.  This choice of representation is motivated as follows: (1) it avoids splitting multi-word concepts (such as \ex{degenerative disease of the central nervous system}) into less meaningful units (\ex{of}, \ex{the}, \ex{central}, etc.), as would happen in the case of token-based representation; and (2) it maps different lexico-grammatical variations of a given term into a single concept, thus reducing noise in the data, and highlighting important keywords. For example, concept C0000731 occurs in the articles as \ex{abdominal distension}, \ex{abdominal distention}, \ex{bloating}, \ex{distended abdomens}, \ex{swelling of abdomen}, etc., which would not be captured by typical approaches to text normalisation such as lemmatisation, stemming, or $n$-gram overlap. We represent each MetaMap-disambiguated concept by the lexicalisation that occurs most frequently in the document collection, as distinct from the MetaMap ``preferred term'', which is often a technical description rather than its lexical form (e.g.\ we use \ex{colon} instead of preferred term \ex{Colon structure (body structure)}). 
As the last step, we remove stopwords based on PubMed's stopword list\footnote{\tiny{\href{https://www.ncbi.nlm.nih.gov/books/NBK3827/table/pubmedhelp.T.stopwords/}{ncbi.nlm.nih.gov/books/NBK3827/table/\\pubmedhelp.T.stopwords/}}} and the 100 most frequent tokens in the dataset. The resulting representation (Table~\ref{tab:concepts}) forms the foundation of the remaining analysis and visualisations.

\begin{table*}[t]
\centering
\small
\renewcommand{\arraystretch}{1.5}
\begin{tabulary}{\linewidth}{LL}
  \toprule
    \textbf{Abstract}      & \ex{porcine deltacoronavirus (pdcov) causes severe diarrhea and vomiting in affected piglets. the aim of this study was to establish the basic, in vitro characteristics of the life cycle such as replication kinetics, cellular ultrastructure, virion morphology, and induction of autophagy of pdcov.}                                                 \\
    
    \textbf{Concept IDs}       & C4038448 C1314792 C1443924 C0042963 C0392760 C2948600 C0557651 C0040363 C0443211 C1527178 C2827718 C1521970 C0023675 C0598312 C0022702 C0007634 C0041623 C0042760 C0332437 C0857127 C0004391  \\
    
    \textbf{Concepts} & porcine deltacoronavirus, causes, severe diarrhea, vomiting, affected, aim, study, to, establish, based, in vitro, characteristics, life cycle, replication, kinetics, cells, ultrastructural, virion, morphology, induction, autophagy \\
    
    \textbf{Final terms} & porcine deltacoronavirus, causes, severe diarrhea, vomiting, affected, aim, establish, in vitro, characteristics, life cycle, replication, kinetics, ultrastructural, virion, morphology, induction, autophagy \\
  \bottomrule
\end{tabulary}
\caption{Text representation steps}
\label{tab:concepts}
\end{table*}

\subsection{Topic view}

To enable document exploration based on common semantic themes, we provide a topic modelling view. We use latent Dirichlet allocation (LDA) \citep{blei2003latent} to learn topics over the whole dataset, and represent the topic mixtures for a given subset of articles (based on the briefcase) as a 2-dimensional map using principal component analysis (PCA) via the pyLDAvis package.\footnote{\tiny{\href{https://pypi.org/project/pyLDAvis/}{pypi.org/project/pyLDAvis}}} LDA represents each document as a mixture of topics, and each topic as a mixture of words. We choose the optimal number of topics (20) based on the $C_v$ topic coherence measure \citep{roder2015exploring}. 
After learning the topic distribution for the whole dataset, we display only those topics which are representative of the selected subset, that is, ones for which the average weight across the selected documents is over a threshold \textit{t}, which we set to 0.05. Thus, the topic selection is dynamic for each set of selected documents.

\subsection{Concept cloud view}

For each of the documents in the user's briefcase, we display a wordcloud\footnote{\tiny{\href{https://github.com/chrisrzhou/react-wordcloud}{github.com/chrisrzhou/react-wordcloud}}} containing the 20 most representative concepts. We regard each of the articles in the briefcase as a target corpus, and the remainder of the articles in the briefcase as a background corpus, and compare concept distributions using the log-likehood test \citep{rayson2000comparing}. This highlights concepts that differentiate a particular document from others in the briefcase, even if they discuss similar topics and have the same set of frequent terms. For instance, if the briefcase contains articles regarding mechanical ventilation and we choose one of them,\footnote{\tiny{\href{https://pubmed.ncbi.nlm.nih.gov/18440440/}{pubmed.ncbi.nlm.nih.gov/18440440}}} the top 10 concepts chosen by log-likelihood will be similar to 10 most frequent concepts, but with two visible differences (Table~\ref{tab:salient}): the log-likelihood results do not contain common terms as \ex{mechanical ventilation}, frequently used in all documents in the briefcase; on the other hand, they include more document-specific concepts as \ex{ai} (\ex{avian influenza)}).
Table~\ref{tab:salient} also shows how operating at the concept level captures multi-word terms such as \ex{mechanical ventilation} and \ex{admitted to intensive care unit}.

\begin{table*}[t]
\centering
\small
\renewcommand{\arraystretch}{1.5}
\begin{tabulary}{\textwidth}{LJJ}
  \toprule
     & Concepts & Tokens
    \\
     \cmidrule(l){1-3}

Frequent & sars-cov, airborne, mechanical ventilation, people, countries, killed, required, medical, community, threat & mechanical, airborne, ventilation, respiratory, people, countries, killed, required, intensive, care \\
Log-likelihood & epidemic, sars-cov, airborne, people, killed, community, threat, ai, countries, required, admitted to intensive care unit & airborne, people, killed, infectious, diseases, caused, community, ai, will, countries \\
  \bottomrule
\end{tabulary}
\caption{Representative concepts vs.\ tokens}
\label{tab:salient}
\end{table*}

\subsection{Technical details}

All data is stored in a graph database (neo4j\footnote{\tiny{\href{https://neo4j.com/}{neo4j.com}}}). The front-end of our web application interacts with the database via the Cypher language and the py2neo\footnote{\tiny{\href{http://py2neo.org/}{py2neo.org}}} library.
The website was built with React\footnote{\tiny{\href{https://reactjs.org/}{reactjs.org}}} and Flask\footnote{\tiny{\href{https://flask.palletsprojects.com/}{flask.palletsprojects.com}}}, and topic visualisations are supported by 
pyLDAVis\footnote{\tiny{\href{https://github.com/bmabey/pyLDAvis}{github.com/bmabey/pyLDAvis}}}. 
Core analysis was done in Python.

\section{Related work}
In the first months of the COVID-19 crisis, many tools were released to support literature exploration. We present here the systems that, like ours, focus on the visualisation of concepts and relations for retrieved documents, or that use concept information to guide search. We provide a summary of the existing tools and their functionalities in Table~\ref{tab:approaches}. 

\textbf{AllenAI's SciSight} Faceted Search \citep{Hope2020SciSightCF}\footnote{\tiny{\href{https://scisight.apps.allenai.org/}{scisight.apps.allenai.org}}} is a tool for exploring how authors and topics interact over time. The user can select the desired topics from different PICO-like categories, which act as a filter for the shown articles. 
It also provides 
a co-mentions view with chord diagrams, which displays associations between diseases and chemicals, or between proteins, genes and cells. This is conceptually similar to our goal of representing concept relations, but their diagram view does not support 
PICO.

The identification of PICO elements in SciSight is based on \textbf{DOC Search},\footnote{\tiny{\href{https://covid-search.doctorevidence.com/}{covid-search.doctorevidence.com}}},  
where a user constructs a query by accepting auto-suggestions based on an ontology. The selected terms can be combined with a variety of Boolean and ontological 
operators. The user can then use the data visualisations to further narrow down the search results. DOC Search represents the document collection with different views, but does not display  topics or relations between PICO concepts, and 
does not support 
natural language queries. That said, DOC Search is a powerful tool for navigating biomedical research publications. 

\textbf{IBM COVID-19 Navigator}\footnote{\tiny{\href{https://covid-19-navigator.mybluemix.net/}{covid-19-navigator.mybluemix.net}}} is similar in functionality to DOC Search. 
The tool fully embraces the use of the UMLS knowledge graph to support semantic search, with Boolean search operators and UMLS semantic relationships. It does not provide  PICO or 
visualisation functionality. 

\textbf{COVID-19 LOVE}\footnote{\tiny{\href{https://app.iloveevidence.com/loves/5e6fdb9669c00e4ac072701d?utm=aile}{app.iloveevidence.com/loves/5e6fdb9669c00e4ac072701d}}} from the Epistemonikos project organises the literature according to study type (systematic reviews, broad syntheses, primary studies), as well as PICO categories and question types (e.g.\ diagnosis, prognosis), but is not based on automatic analysis or on the CORD collection, and does not provide visual exploration.

\textbf{Trialstreamer} \citep{nye-etal-2020-trialstreamer}
finds and summarises new clinical trial publications, registrations, and preprints in both COVID-19-related and broader literature. PICO snippets of text are shown, together with the key findings and an indication of the risk of bias. 
\textit{Evidence maps} link interventions 
to outcomes, and predict a trial's findings.

Other systems include: \textbf{NIH's LitCovid} \citep{LitCOVID}\footnote{\tiny{\href{https://www.ncbi.nlm.nih.gov/research/coronavirus/}{ncbi.nlm.nih.gov/research/coronavirus}}}, which provides---based on human verification---categorisation of papers according to diagnosis, treatment, etc.; 
\textbf{WellAI}'s tool\footnote{\tiny{\href{https://wellai.health/covid/}{wellai.health/covid}}} for concept search, enabling retrieval of documents via highly associated concepts;
\textbf{SemViz} \citep{tu2020exploration}, 
using a knowledge graph from Blender Lab\footnote{\tiny{\href{http://blender.cs.illinois.edu/covid19/}{blender.cs.illinois.edu/covid19}}} to create interaction plots and word clouds for chemicals, genes and diseases;  and \textbf{COVID Intelligent Search}\footnote{\tiny{\href{https://covidsearch.sinequa.com/}{covidsearch.sinequa.com}}}, which offers filters for various medical categories (e.g.\ drugs, 
indication), and suggests textual extracts as queries. \citet{bras2020visualising} present work on coarse-to-fine exploration of literature based on hierarchical topic models, however not 
 visualisation of arbitrary sets of articles. \citet{ahamed2020information} construct a graph from  entity co-occurrences 
 that enables centrality-based ranking of drugs, pathogens and biomolecules.

\begin{table}[t]
\addtolength{\tabcolsep}{-3.5pt} 

\centering
\small
\begin{tabular}{l l l l l l l l l l l l}
  \toprule
& \rot{COVID-SEE (ours)} & \rot{SciSight} & \rot{DOC Search} & \rot{COVID-19 Navigator} &\rot{LitCOVID} & \rot{SemViz} & \rot{WellAI} &\rot{COVID Intel.\ Search} & \rot{Le Bras et al.\ 2020} & \rot{COVID-19 LOVE} & \rot{Trialstreamer} \\
\cmidrule(l){1-12}
\textbf{Search:} & & & & & & & & & & &\\
\hspace{0.1cm} NL/IR & \cmark & \xmark
& \cmark & \xmark  & \cmark & \xmark & \xmark & \cmark & \xmark & \xmark & \cmark \\
\hspace{0.1cm} Concepts & \xmark & \cmark & \cmark & \cmark & \cmark & \cmark & \cmark & \cmark & \cmark
& \cmark & \cmark \\
\hspace{0.1cm} PICO & \xmark
& \cmark & \cmark & \xmark & \xmark & \xmark & \xmark & \xmark &\xmark & \cmark & \cmark \\
\textbf{Visualisation:} & & & & & & & & & & & \\
\hspace{0.1cm} Concepts & \cmark & \textbf{?} & \textbf{?} & \xmark & \xmark & \cmark & \xmark & \xmark & \xmark & \xmark & \textbf{?}\\
\hspace{0.1cm} Relations & \cmark & \cmark & \xmark & \xmark & \xmark & \cmark & \xmark & \xmark & \xmark & \xmark & \xmark
\\
\hspace{0.1cm} Topics & \cmark & \xmark & \xmark & \xmark & \xmark & \xmark & \xmark & \xmark & \textbf{?} & \xmark & \xmark \\
  \bottomrule
\end{tabular}
\caption{Comparison with related systems.}
\label{tab:approaches}
\end{table}


\section{Limitations and future work}
COVID-SEE has been designed to facilitate more interactive exploration of the COVID-19 literature, through integration of sub-collection thematic analysis, document-level visual concept summaries, and PICO-structured concept relations. 
It does not currently support semantic search based on PICO-style queries; a natural extension would be to add this functionality 
utilising the PICO and UMLS concept pre-processing.
A recommendation system for articles which have similar topic distributions could be added. 
In terms of visual representations, we will experiment with expanding beyond the MeSH term vocabulary to include more specific terminology, and
use of the hierarchical relationships that exist between terms. Finally, we are planning a user study with medical professionals 
to evaluate the potential of COVID-SEE as a knowledge discovery tool. 


\bibliographystyle{acl_natbib}
\bibliography{anthology,emnlp2020}

\begin{thebibliography}{22}
\expandafter\ifx\csname natexlab\endcsname\relax\def\natexlab#1{#1}\fi

\bibitem[{Ahamed and Samad(2020)}]{ahamed2020information}
Sabber Ahamed and Manar Samad. 2020.
\newblock {Information Mining for COVID-19 Research From a Large Volume of
  Scientific Literature}.
\newblock \emph{arXiv:2004.02085}.

\bibitem[{Blei et~al.(2003)Blei, Ng, and Jordan}]{blei2003latent}
David~M Blei, Andrew~Y Ng, and Michael~I Jordan. 2003.
\newblock {Latent Dirichlet allocation}.
\newblock \emph{Journal of Machine Learning Research}, 3(Jan):993--1022.

\bibitem[{Chen et~al.(2020)Chen, Allot, and Lu}]{LitCOVID}
Q.~Chen, A.~Allot, and Z.~Lu. 2020.
\newblock \href {https://doi.org/10.1038/d41586-020-00694-1} {Keep up with the
  latest coronavirus research}.
\newblock \emph{Nature}, 579(7798):193.

\bibitem[{G{\"o}rg et~al.(2010)G{\"o}rg, Tipney, Verspoor, Baumgartner, Cohen,
  Stasko, and Hunter}]{gorg2010visualization}
Carsten G{\"o}rg, Hannah Tipney, Karin Verspoor, William~A Baumgartner,
  K~Bretonnel Cohen, John Stasko, and Lawrence~E Hunter. 2010.
\newblock Visualization and language processing for supporting analysis across
  the biomedical literature.
\newblock \emph{International Conference on Knowledge-Based and Intelligent
  Information and Engineering Systems}, pages 420--429.

\bibitem[{Hakala et~al.(2016)Hakala, Kaewphan, Salakoski, and
  Ginter}]{hakala-etal-2016-syntactic}
Kai Hakala, Suwisa Kaewphan, Tapio Salakoski, and Filip Ginter. 2016.
\newblock \href {https://doi.org/10.18653/v1/W16-2913} {Syntactic analyses and
  named entity recognition for {P}ub{M}ed and {P}ub{M}ed central {---}
  up-to-the-minute}.
\newblock In \emph{Proceedings of the 15th Workshop on Biomedical Natural
  Language Processing}, pages 102--107, Berlin, Germany. Association for
  Computational Linguistics.

\bibitem[{Hope et~al.(2020)Hope, Portenoy, Vasan, Borchardt, Horvitz, Weld,
  Hearst, and West}]{Hope2020SciSightCF}
Tom Hope, Jason Portenoy, Kishore Vasan, Jonathan Borchardt, Eric Horvitz,
  Daniel~S. Weld, Marti~A. Hearst, and Jevin~D. West. 2020.
\newblock {SciSight: Combining faceted navigation and research group detection
  for COVID-19 exploratory scientific search}.
\newblock \emph{bioRxiv}.

\bibitem[{Hutson(2020)}]{Hutson2020}
Matthew Hutson. 2020.
\newblock \href {https://doi.org/10.1038/d41586-020-01733-7}
  {Artificial-intelligence tools aim to tame the coronavirus literature}.
\newblock \emph{Nature}.

\bibitem[{Lample et~al.(2016)Lample, Ballesteros, Subramanian, Kawakami, and
  Dyer}]{lample-etal-2016-neural}
Guillaume Lample, Miguel Ballesteros, Sandeep Subramanian, Kazuya Kawakami, and
  Chris Dyer. 2016.
\newblock \href {https://doi.org/10.18653/v1/N16-1030} {Neural architectures
  for named entity recognition}.
\newblock In \emph{Proceedings of the 2016 Conference of the North {A}merican
  Chapter of the Association for Computational Linguistics: Human Language
  Technologies}, pages 260--270, San Diego, California. Association for
  Computational Linguistics.

\bibitem[{{Le Bras} et~al.(2020){Le Bras}, Gharavi, Robb, Vidal, Padilla, and
  Chantler}]{bras2020visualising}
Pierre {Le Bras}, Azimeh Gharavi, David~A. Robb, Ana~F. Vidal, Stefano Padilla,
  and Mike~J. Chantler. 2020.
\newblock {Visualising COVID-19 Research}.
\newblock \emph{arXiv:2005.06380}.

\bibitem[{Lindberg et~al.(1993)Lindberg, Humphreys, and
  McCray}]{lindberg1993unified}
Donald~AB Lindberg, Betsy~L Humphreys, and Alexa~T McCray. 1993.
\newblock The {Unified Medical Language System}.
\newblock \emph{Yearbook of Medical Informatics}, 2(01):41--51.

\bibitem[{Marchionini(2006)}]{marchioni2006}
Gary Marchionini. 2006.
\newblock \href {https://doi.org/10.1145/1121949.1121979} {Exploratory search:
  From finding to understanding}.
\newblock \emph{Commun. ACM}, 49(4):41–46.

\bibitem[{Nye et~al.(2018)Nye, Li, Patel, Yang, Marshall, Nenkova, and
  Wallace}]{nye-etal-2018-corpus}
Benjamin Nye, Junyi~Jessy Li, Roma Patel, Yinfei Yang, Iain Marshall, Ani
  Nenkova, and Byron Wallace. 2018.
\newblock \href {https://doi.org/10.18653/v1/P18-1019} {A corpus with
  multi-level annotations of patients, interventions and outcomes to support
  language processing for medical literature}.
\newblock In \emph{Proceedings of the 56th Annual Meeting of the Association
  for Computational Linguistics (Volume 1: Long Papers)}, pages 197--207,
  Melbourne, Australia. Association for Computational Linguistics.

\bibitem[{Nye et~al.(2020)Nye, Nenkova, Marshall, and
  Wallace}]{nye-etal-2020-trialstreamer}
Benjamin Nye, Ani Nenkova, Iain Marshall, and Byron~C. Wallace. 2020.
\newblock \href {https://www.aclweb.org/anthology/2020.acl-demos.9}
  {{T}rialstreamer: Mapping and browsing medical evidence in real-time}.
\newblock In \emph{Proceedings of the 58th Annual Meeting of the Association
  for Computational Linguistics: System Demonstrations}, pages 63--69, Online.
  Association for Computational Linguistics.

\bibitem[{Pang et~al.(2015)Pang, Verspoor, Chang, and Pearce}]{pang2015}
Patrick Cheong-Iao Pang, Karin Verspoor, Shanton Chang, and Jon Pearce. 2015.
\newblock \href {https://doi.org/10.1007/s12553-015-0096-0} {Conceptualising
  health information seeking behaviours and exploratory search: result of a
  qualitative study}.
\newblock \emph{Health and Technology}, 5(1):45--55.

\bibitem[{Rayson and Garside(2000)}]{rayson2000comparing}
Paul Rayson and Roger Garside. 2000.
\newblock Comparing corpora using frequency profiling.
\newblock \emph{The workshop on comparing corpora}, pages 1--6.

\bibitem[{Richardson et~al.(1995)Richardson, Wilson, Nishikawa, Hayward
  et~al.}]{richardson1995well}
W~Scott Richardson, Mark~C Wilson, Jim Nishikawa, Robert~S Hayward, et~al.
  1995.
\newblock The well-built clinical question: a key to evidence-based decisions.
\newblock \emph{Acp j club}, 123(3):A12--3.

\bibitem[{Roberts et~al.(2020)Roberts, Alam, Bedrick, Demner-Fushman, Lo,
  Soboroff, Voorhees, Wang, and Hersh}]{roberts-etal-2020-trec}
Kirk Roberts, Tasmeer Alam, Steven Bedrick, Dina Demner-Fushman, Kyle Lo, Ian
  Soboroff, Ellen Voorhees, Lucy~Lu Wang, and William~R Hersh. 2020.
\newblock \href {https://doi.org/10.1093/jamia/ocaa091} {{TREC-COVID: Rationale
  and Structure of an Information Retrieval Shared Task for COVID-19}}.
\newblock \emph{Journal of the American Medical Informatics Association}.
\newblock Ocaa091.

\bibitem[{R{\"o}der et~al.(2015)R{\"o}der, Both, and
  Hinneburg}]{roder2015exploring}
Michael R{\"o}der, Andreas Both, and Alexander Hinneburg. 2015.
\newblock Exploring the space of topic coherence measures.
\newblock \emph{Proceedings of the Eighth ACM International Conference on Web
  Search and Data Mining}, pages 399--408.

\bibitem[{Stasko et~al.(2008)Stasko, Görg, and Liu}]{stasko2008}
John Stasko, Carsten Görg, and Zhicheng Liu. 2008.
\newblock \href {https://doi.org/10.1057/palgrave.ivs.9500180} {Jigsaw:
  Supporting investigative analysis through interactive visualization}.
\newblock \emph{Information Visualization}, 7(2):118--132.

\bibitem[{Tu et~al.(2020)Tu, Verhagen, Cochran, and
  Pustejovsky}]{tu2020exploration}
Jingxuan Tu, Marc Verhagen, Brent Cochran, and James Pustejovsky. 2020.
\newblock {Exploration and Discovery of the COVID-19 Literature through
  Semantic Visualization}.
\newblock \emph{arXiv:2007.01800}.

\bibitem[{Wang et~al.(2020)Wang, Lo, Chandrasekhar, Reas, Yang, Eide, Funk,
  Kinney, Liu, Merrill, Mooney, Murdick, Rishi, Sheehan, Shen, Stilson, Wade,
  Wang, Wilhelm, Xie, Raymond, Weld, Etzioni, and
  Kohlmeier}]{wang-lo-2020-cord19}
Lucy~Lu Wang, Kyle Lo, Yoganand Chandrasekhar, Russell Reas, Jiangjiang Yang,
  Darrin Eide, Kathryn Funk, Rodney Kinney, Ziyang Liu, William Merrill, Paul
  Mooney, Dewey Murdick, Devvret Rishi, Jerry Sheehan, Zhihong Shen, Brandon
  Stilson, Alex~D. Wade, Kuansan Wang, Chris Wilhelm, Boya Xie, Douglas
  Raymond, Daniel~S. Weld, Oren Etzioni, and Sebastian Kohlmeier. 2020.
\newblock \href {arXiv:2004.10706} {{CORD-19: The Covid-19 Open Research
  Dataset}}.
\newblock In \emph{Proceedings of the Workshop on {NLP} for {COVID-19} at {ACL
  2020}}, Online. Association for Computational Linguistics.

\bibitem[{Zhang et~al.(2020)Zhang, Gupta, Nogueira, Cho, and
  Lin}]{zhang2020rapidly}
Edwin Zhang, Nikhil Gupta, Rodrigo Nogueira, Kyunghyun Cho, and Jimmy Lin.
  2020.
\newblock {Rapidly Deploying a Neural Search Engine for the COVID-19 Open
  Research Dataset: Preliminary Thoughts and Lessons Learned}.
\newblock arXiv:2004.05125.

\end{thebibliography}


\end{document}